\documentclass[preprint,12pt,3p]{elsarticle}
\makeatletter
\def\ps@pprintTitle{%
 \let\@oddhead\@empty
 \let\@evenhead\@empty
 \def\@oddfoot{\centerline{\thepage}}%
 \let\@evenfoot\@oddfoot}
\makeatother

\usepackage{amssymb}

\usepackage{amsmath}
\usepackage{amssymb}
\usepackage{ragged2e}
\usepackage{amsmath}
\usepackage[linesnumbered,ruled]{algorithm2e} 

\usepackage{graphicx}
\usepackage{url}
\usepackage{natbib}

\begin{document}

\begin{frontmatter}

\title{Bayesian active learning for choice models with deep Gaussian processes}

\author{Jie Yang$^a$}
\address[1]{Department of Civil and Environmental Engineering, Northwestern University, Evanston, IL, USA}


\ead{jieyang2011@u.northwestern.edu}

\author{Diego Klabjan$^b$}
\address[2]{Department of Industrial Engineering and Management Sciences, Northwestern University, Evanston, IL, USA}
\ead{d-klabjan@northwestern.edu }

\begin{abstract}
In this paper, we propose an active learning algorithm and models which can gradually learn individual's preference through pairwise comparisons. The active learning scheme aims at finding individual's most preferred choice with minimized number of pairwise comparisons. The pairwise comparisons are encoded into probabilistic models based on assumptions of choice models and deep Gaussian processes. The next-to-compare decision is determined by a novel acquisition function. We benchmark the proposed algorithm and models using functions with multiple local optima and one public airline itinerary dataset. The experiments indicate the effectiveness of our active learning algorithm and models.

\end{abstract}


\end{frontmatter}

\setlength\parindent{0pt}

\section{Introduction}

The airline industry is evolving. After years of consolidation, carriers are refocusing their competitive advantages on promoting and merchandising the right product to the right traveler at the right time. To gather more valuable travelers' data and squeeze the distribution structure, carriers are encouraging travelers to purchase through their own website or app. IATA (Harteveldt, 2016) projects purchase through airline mobile app will surge from 2\% to 7\% while website purchase will increase from 33\% to 37\% by year 2021. This creates great opportunities for carriers to achieve traveler personalization and increase their travelers' stickiness as currently rich travelers' information is retained by distributors. Personalization also means traditional methods of using aggregate data to infer travelers' preference may no longer be satisfactory and techniques requiring smaller samples from the same individual may be more beneficial.

\vspace{5mm}
An airline capturing travelers' online actions related to promotions such as click-throughs or bookings sits on a trove of data. These online actions may result in clicked offers, purchased offers and non-clicked offers. Furthermore, the airline can record more information such as visit duration based on an offer, number of repeated clicks based on a different offer, etc. With the collected actions and information, the airline can construct pairwise comparisons each time it promotes. On the next occasion, more attractive promotions can be actively found through an active learning algorithm and pushed to the traveler based on learned preference. One advantage of active learning is a low number of interactions with the traveler in order to learn his or her preferences. This not only improves customer experience but also increases revenue. 

\vspace{5mm}
We develop an algorithm and model that can actively search for one individual's most preferred choice based exclusively on pairwise comparisons. We rely on an individual's preference over different choices as a latent function depending on choices' features. The probabilistic model takes pairwise comparisons and latent function values as input. We apply pairwise comparison as it is a more effective approach to collect individual's preference than using a choice set with multiple choices. With a series of pairwise comparisons as the input, we approximate the joint distribution describing the comparisons first and then build a probabilistic model which is different from multiplying each pairwise probability one by one under the traditional assumption of independency. In order to capture discrete choice models our noise is Gumbel distributed and not normal as standardly assumed. This yields the fact that the joint distribution over pairs does not decompose. The goal of the overall algorithm is to minimize the number of queries we request before finding the most preferred choice. One difference between our active learning algorithm and the past works is that we use deep Gaussian process (DGP) priors which have been studied in other supervised tasks (Damianou and Lawrence, 2013; Bui et al., 2016). In the airline industry and elsewhere, choice models are commonly used in stated preference inference. Within that, Gumbel noises are a required assumption and this motivates us to develop models using similar assumptions. In the following sections, we use ``instance,'' ``choice'' and ``itinerary'' interchangeably. 

\vspace{5mm}
The contributions of this paper are as follows:
\begin{itemize}
	\item To the best of our knowledge, we are the first to use DGPs, not only in conjunction to Gumbel noises but also in preference-based active learning;
	\item We apply correlated Gumbel noises to active preference learning. To our best knowledge, no prior works have focused on this since normal noise is always assumed. This captures the setting of nested logit models popular in discrete choice utility studies;
    \item Deviating from traditional choice models which use a linear combination between weights and features as the utility, we assume DGP priors to the latent utility;
    \item We develop a way to approximate a preference chain derived from pairwise comparisons which proved to be effective by means of the experiments;
    \item A new acquisition function has been proposed and used in the active learning process.
\end{itemize}

\vspace{5mm}
The rest of the paper is organized as follows. Section 2 provides the literature review and Section 3 introduces the preliminaries needed. Section 4 provides model assumptions, the DGP models, probabilistic formulations, the acquisition function and the active learning algorithm. Section 5 includes experiments on latent functions and an airline itinerary dataset.

\section{Literature review}
\vspace{5mm}

Travel demand and preference studies have previously focused on choice models with wide use in airlines' revenue management systems such as passenger origin-destination simulators (Carrier and Weatherford, 2015). For empirical studies using choice models, we refer to Coldren and Koppelman (2005), Warburg et al. (2006) and Garrow (2010). In these studies, multinomial logit models, nested logit (NL) models, mixed logit models and generalized extreme value (GEV) models are widely used. For details of these models including model assumptions and formulas, we refer to McFadden (1978), Ben-Akiva and Bierlaire (1999) and Koppelman and Sethi (2000).

\vspace{5mm}

We rely on Bayesian optimization which is applicable in situations where a closed-form expression of the objective function is not available, but where observations (possibly noisy) of this function at sampled values (Brochu et al., 2010) are readily available. Instead of modeling the utility function as a linear combination between weights and features which is a traditional practice in choice models, we assume utilities over different choices follow a black box function. Different from Bayesian optimization which returns the function value at each step, our model learns hidden values from pairwise preference relations using a probabilistic approach. In this way, the algorithm only learns which choice/point yields a higher utility/function value. Chu and Ghahramani (2005) introduce a nonparametric Bayesian approach to preference learning over instances or labels with the latent function values imposed by GP priors. Brochu et al. (2007) develop an algorithm based on GP which automatically decides what items are best presented to an individual in order to find the item that they value highly in as few trials as possible. The difference between Brochu et al. (2007) and Chu and Ghahramani (2005) is that the latter selects instances from a finite pool while Brochu et al. (2007) maximize a latent function over a continuous space. A random selection method is used in Brochu et al. (2007) to search a continuous space and the results are used as a baseline which is a common benchmark in active learning for regression models. Our work is different from previous works in three ways. Firstly, instead of using probit or Thurstone-Mosteller models, our assumptions are based on NL models and this leads to more complex probabilistic formulations but more realistic models. Secondly, besides GP priors which can be considered as simplified DGPs with no hidden hierarchies, we mainly focus on DGP models that in turn yield better performances. Thirdly, the proposed modeling assumptions lead to a unique probability distribution to compare instances. We approximate this distribution and incorporate it in the associated acquisition function within the active learning algorithm.
\vspace{5mm}

GP models have been widely studied in regression and classification. Analytic inference, expressivity, integration over hypotheses and closed-form predictive distribution are the main useful properties of GPs (Duvenaud, 2014). Besides the works on active preference learning, GP models can also be applied to semi-supervised learning. For example, semi-supervised GP (Sindhwani et al., 2005) provides a novel prior that exploits the localized spatial structure spanned by the given unlabeled data. For details of GPs and machine learning, we refer to William and Rasmussen (1996).

\vspace{5mm}
Empirically, deep models seem to have structural advantages that can improve the quality of learning in complicated datasets associated with abstract information (Bengio, 2009). The GP latent variable model (GP-LVM) introduced by Lawrence (2004, 2005) assumes a single hidden layer and treats the hidden layer as latent variables. Titsias and Lawrence (2010) variationally integrate out the latent variables in GP-LVM and compute a closed-form Jensen’s lower bound on the true log marginal likelihood of the data. The key is to expand the probability space with extra variables which allows for priors on the latent space to be propagated through a nonlinear mapping. These extra variables are called inducing points. The concept of using inducing points was originally proposed to solve GP models in Snelson and Ghahramani (2006) with a large dataset because solving the models involves computing the inverse of an $n \times n$ matrix where $n$ is the number of instances. The inducing variables are latent function values evaluated at some subset of the dataset and can be sampled from training data (Titsias, 2009).

\vspace{5mm}

Compared to GP  models or the GP-LVM model, DGP models are more general that may provide better prediction performances. DGP models come with deep structures and use GPs as the mappings to connect layers. A single layer DGP model is effectively the GP-LVM model (Damianou and Lawrence, 2013). Damianou and Lawrence (2013) apply a variational approach which approximately marginalises out the latent space, thus allowing for automatic structure discovery in DGP models' hierarchy. However, in the variational approach (Damianou and Lawrence, 2013), the number of variational parameters increases linearly with the number of training data points which hinders the use of this method for large-scale datasets (Bui et al., 2016). Bui et al. (2016) then use a novel approximate Expectation Propagation (EP) scheme and a probabilistic backpropagation algorithm for DGP models. Different from the variational approach (Damianou and Lawrence, 2013), the algorithm in Bui et al. (2016) does not retain an explicit approximate distribution over the hidden variables. Our algorithmic approach relies on the active learning framework in Brochu et al. (2007) and the DGP approximation in Bui et al. (2016).

\section{Preliminaries}
\label{sec1}
In this section, we start with NL models and then introduce some previous works on active preference learning with probit models, and an approximation in DGP models. Suppose we have a set of choice vectors $X=\{x_1,x_2,...,x_n\}$ with $x_i \in \mathbb{R}^p$, and a set of pairwise preference relations at step $t$, $D_t=\{x_i \succ x_j, x_i,x_j\in X\}_{ij}$. The active learning algorithm adds a new pairwise comparison to $D_t$ after completing step $t$. We assume there is an unobservable latent function value $u(x_i)$ for each $x_i$ so the latent values $\{u(x_1),u(x_2),...u(x_n)\}$ reflect pairwise comparisons in $D_t$ well, i.e. if $x_i \succ x_j$, then $u(x_i)+\epsilon_i \geq u(x_j)+\epsilon_j$ where $\epsilon_i$ and $\epsilon_j$ are noises. The goal is to find $\max_{i} u(x_i)$ based on $D_t$. 

\subsection{NL models}

NL models belong to the family of GEV models. In GEV models, the unobservable portions of utility for all alternatives are jointly distributed as generalized extreme values. In general, NL models assume that a set of choices can be grouped into a nest. For example, non-stop itineraries can be grouped into one nest while one-stop itineraries belong to another nest. If we have a set of unobserved utilities $\{\epsilon_1^u,\epsilon_2^u,...,\epsilon_n^u\}$ for instances in $X$, then the cumulative joint distribution is
\begin{equation}
\exp(-\sum_{m=1}^{M}{(\sum_{i \in B_m}{e^{\frac{-\epsilon_i^u}{\lambda_m}}})^{\lambda_m}}).
\end{equation}
In equation (1), $m$ is the nest index and $B_m$ refers to the set of all choices within nest $m$.
Value $\lambda_m$ can be used to measure the correlation between unobserved utilities among choices within the same nest $m$. If $U_k$ is the utility of choice $k$, then the correlation between $U_i$ and $U_j$ within the same nest $m$ can be written as
\begin{equation*}
corr(U_i,U_j)=1-\lambda_m^2.
\end{equation*}

\vspace{5mm}
Based on assumptions on NL models, we next summarize the expressions of $\mathcal{P}(x_i \succ x_j)$ and $\mathcal{P}(x_i \succ x_j \succ x_k)$ (Dagsvik and Liu, 2006; Oviedo and Yoo, 2016). For $\mathcal{P}(x_i \succ x_j)$, if $x_i$ and $x_j$ are in the same nest $m$ then
\begin{equation}
\mathcal{P}(x_i \succ x_j)=\frac{e^{u_i/\lambda_m}}{e^{u_j/\lambda_m}+e^{u_i/\lambda_m}}.
\end{equation}
If  $x_i$ and $x_j$ are in different nests, we have
\begin{equation}
\mathcal{P}(x_i \succ x_j)=\frac{e^{u_i}}{e^{u_j}+e^{u_i}}.
\end{equation}
With regard to $\mathcal{P}(x_i \succ x_j \succ x_k)$, there are 5 cases to consider where $x_i$  is in nest $m$, $x_j$ in nest $m^{'}$ and $x_k$ in nest $\bar{m}$:

\begin{itemize}
	\item Case 1: $m=m^{'}=\bar{m}$
	\item Case 2: $m^{'}=\bar{m} \neq m$
	\item Case 3: $m^{'}\neq\bar{m}, (m\neq m^{'}$ or $m\neq \bar{m}$)
	\item Case 4: $m=m^{'}, m^{'}\neq \bar{m}$
	\item Case 5: $m=\bar{m}, m^{'}\neq \bar{m}$.
\end{itemize}
Corresponding to these cases, we have the probability formulas defined in Table 1.
\vspace{5mm}
\begin{center}
Table 1. Probability expressions
\end{center}
\begin{center}
\begin {tabular}{l | r}
\hline
Case \#		& Probability expression \\
\hline
\\
Case 1 		& $\frac{e^{u_j/\lambda_m}}{e^{u_j/\lambda_m}+e^{u_k/\lambda_m}}-\frac{e^{u_j/\lambda_m}}{e^{u_i/\lambda_m}+e^{u_j/\lambda_m}+e^{u_k/\lambda_m}}$\\
\\
Case 2          & $\frac{e^{u_j/\lambda_{m^{'}}}}{e^{u_j/\lambda_{m^{'}}}+e^{u_k/\lambda_{m^{'}}}}-\frac{e^{u_j/\lambda_{m^{'}}}(e^{u_j/\lambda_{m^{'}}}+e^{u_k/\lambda_{m^{'}}})^{\lambda_{m^{'}}-1}}{e^{u_i}+(e^{u_j/\lambda_{m^{'}}}+e^{u_k/\lambda_{m^{'}}})^{\lambda_{m^{'}}}}$\\
\\
Case 3            & $\frac{e^{u_j}}{e^{u_j}+e^{u_k}}-\frac{e^{u_j}}{e^{u_i}+e^{u_j}+e^{u_k}}$\\
\\
Case 4              & $\frac{e^{u_j}}{e^{u_j}+e^{u_k}}-\frac{e^{u_j/\lambda_m}(e^{u_j/\lambda_m}+e^{u_i/\lambda_m})^{\lambda_m-1}}{e^{u_k}+(e^{u_j/\lambda_m}+e^{u_i/\lambda_m})^{\lambda_m}}$\\
\\
Case 5 		& $\frac{e^{u_j}}{e^{u_j}+e^{u_k}}-\frac{e^{u_j}}{e^{u_j}+(e^{u_i/\lambda_m}+e^{u_k/\lambda_m})^{\lambda_m}}$\\
&\\
\hline
\end {tabular}
\end{center}
\vspace{5mm}

\subsection{Active Preference Learning with Probit Models}

Here we summarize the approach from Brochu et al. (2007). In active preference learning, the objective is
\begin{equation*}
	max_{x}\ u(x|D_{t-1})+\epsilon.
\end{equation*}
When using probit models, $\epsilon$ is assumed to be normally distributed with variance $\sigma_{noise}$. Then the probability of a pairwise comparison can be developed as $\mathcal{P}(x_i \succ x_j) = \Phi(\frac{u(x_i)-u(x_j)}{\sqrt{2}\sigma_{noise}})$ where $\Phi()$ is the cumulative density function of the standard normal distribution.

\vspace{5mm}
The active preference learning approach in Brochu et al. (2007) with probit models assumes GP priors on $u(x_i)$. In GPs, the prior probability of latent values  $\{u(x_i)\}$ is a multivariate Gaussian distribution with
\begin{equation}
\mathcal P(u)=\frac{1}{(2\pi)^{\frac{n}{2}}\left |\Sigma_{u}\right|^{\frac{1}{2}}}\exp(-\frac{1}{2}u^{T}\Sigma_{u}^{-1}u).
\end{equation}
In this distribution, matrix $\Sigma_{u}$ is specified by using Gaussian kernel functions
\begin{equation*}
\mathcal \kappa(x_i,x_j)=\sigma_{SE}^2\exp(-\frac{1}{2l^2}||x_i-x_j||^2).
\end{equation*}
Each $ij^{th}$ element in $\Sigma_{u}$ refers to $\kappa(x_i,x_j)$. 

\vspace{5mm}

For simplicity we use $D$ instead of $D_t$. We next outline the approach from Brochu et al. (2007). To compute or approximate the posterior distribution of $u$ based on $D$, we use Laplace approximation which carries out the Taylor expansion at the maximum a posteriori (MAP) point and retains the terms up to the second order (MacKay, 1994):
\begin{equation*}
\log \mathcal{P}(u|D)  \approx \log \mathcal{P}(u^*|D)+\frac{1}{2}(u-u^*)^{T} \mathcal{H}^*(u-u^*).
\end{equation*}
We denote $u^{*}=arg max_u \mathcal{P}(u|D)$ and $\mathcal{H}^*$ as the Hessian matrix at the MAP point.

\vspace{5mm}

In view of this, we first find hidden values $u$ maximizing $\mathcal{P} (u|D)$. According to Bayes we have $P(u|D) \propto P(D|u)P(u)$ and thus we can find the MAP point by maximizing the probability $\mathcal{P}(D|u)\mathcal{P}(u)$ (Chu and Ghahramani, 2005; Brochu et al., 2010).

\vspace{5mm}
If $\mathcal{L}=\mathcal{P}(D|u)\mathcal{P}(u)$ then
\begin{equation*}
\log \mathcal{L} =\log \mathcal{P}(D|u) + \log \frac{1}{(2\pi)^{\frac{n}{2}}\left |\Sigma_{u}\right|^{\frac{1}{2}}}-\frac{1}{2}u^{T}\Sigma_{u}^{-1}u,
\end{equation*}
since $P(u)$ follows a multivariate Gaussian distribution.
\vspace{5mm}

To get the optimal MAP, we compute the gradient of $\log \mathcal{L}$ as
\begin{equation*}
g(u)=\triangledown_u \log \mathcal{P}(D|u)-\Sigma_{u}^{-1}u.
\end{equation*}
Similarly, we have the Hessian matrix defined as
\begin{equation*}
\mathcal{H}=\triangledown_u \triangledown_u \log \mathcal{P}(D|u)+\Sigma_{u}^{-1}.
\end{equation*}
To find $u^*$, we can use second order optimization methods.

\vspace{5mm}
\subsection{DGP models and EP}

In this outline, we follow Bui et al. (2016). 
In a DGP model,
\begin{equation*}
\begin{split}
U &= u(v),\\
v &= h(x) + \epsilon,
\end{split}
\end{equation*}
with $\epsilon$ as a Gaussian noise and $u(),h()$ as GP mappings. As a result, $\mathcal{P}(u|D)$ is the same as in GP models while $\mathcal{P}(u)$ is different. 
We can compute $\mathcal{P}(u)$ through a marginal distribution by integrating out $v$ as
\begin{equation*}
	\mathcal{P}(u)= \int \mathcal{P}(u|v)\mathcal{P}(v)d_v.
\end{equation*}
\vspace{5mm}

The marginal distribution $\mathcal{P}(u)$ has no explicit formula and is intractable since $v$ and $u$ are treated through different non-linear GP priors. If $\sigma_h$ is the noise at the hidden layer $v$, we can write
\begin{equation*}
\begin{split}
	\mathcal{P}(v|h,X,\sigma_h^2)&= \prod_{i} \mathcal{N}(v_i;h(x_i),\sigma_h^2),\\
	\mathcal{P}(u|v) &= \prod_{i} \mathcal{N}(u_i;u(v_i)).
\end{split}
\end{equation*}

\vspace{5mm}
The flow of the entire propagation is presented in Figure 1. We augment the probability space with $s$ at the hidden and top layers. Each $s$ point has an inducing input, namely $z$. 
Using the bottom layer as an example, $z_1$ and $X$ map through the same GP prior $h()$ to $s_1$ and $h$.

\begin{figure}
 \centering
  \includegraphics[width=130mm]{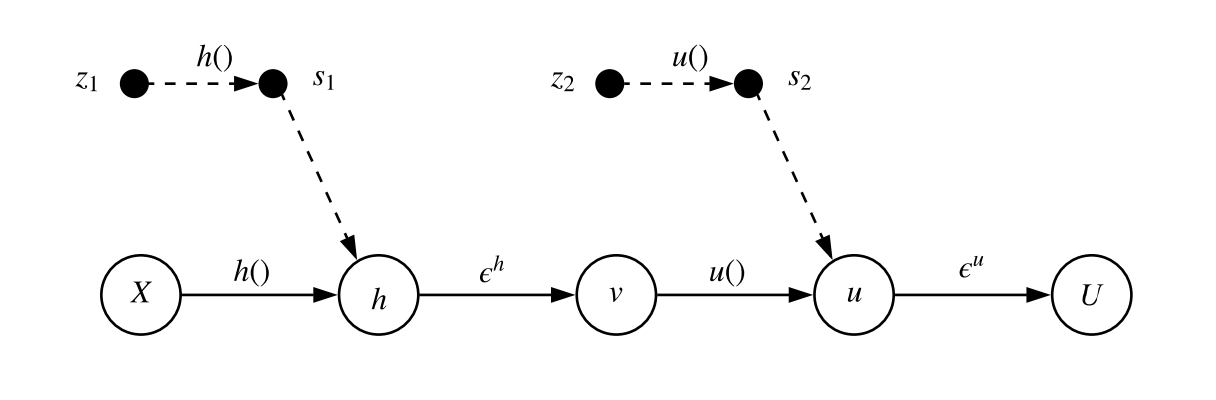}
  \caption{DGP propagation flow}
\end{figure}

\vspace{5mm}
With the inducing points and $X$, we can rewrite and expand the marginal distribution of $\mathcal{P}(u)$ as
\begin{equation*}
	\mathcal{P}(u) = \underbrace{\int (\mathcal{P}(u|v,s_2)\mathcal{P}(s_2)d_{s_2}}_\text{top layer} \underbrace{\int \mathcal{P}(v|X,s_1)\mathcal{P}(s_1)d_{s_1}}_\text{hidden layer})d_{v}.
\end{equation*}

\vspace{5mm}

The EP algorithm provides an approximation for distributions belonging to the exponential family (Seeger, 2007; Bui et al., 2016). Bui et al. (2016) approximate $\log \mathcal{P}(u)$ by
\begin{equation}
	\mathcal{J}(\Theta) = (1-N)\phi(\theta)+N\phi(\theta^{\backslash 1})-\phi(\theta_{prior})+\sum_{i=1}^N \log Z_i.
\end{equation}

Here, $\Theta$ includes all model parameters and $N$ is the number of instances. Vectors $\theta$, $\theta_{prior}$ correspond to natural parameters of the approximate posterior $\mathcal{Q}(s)$ to $\mathcal{P}(s|X,u)$, and prior $\mathcal{P}(s)$, respectively. Vector $\theta^{\backslash 1}$ is the natural parameter of the cavity distribution (not corresponding to a data point). Function $\phi$ is a log normalizer of Gaussian.

\vspace{5mm}

In addition, we have $\log Z_i = \log \int \mathcal{Q}^{\backslash i}(s)\mathcal{P}(u_i|x_i,s) d_{s}$ where $\mathcal{Q}^{\backslash i}(s) \propto \mathcal{Q}(s)/t_i(s)$ with $t_i(s)$ being an approximate data factor. Term $\log Z_i$ can be expanded as
\begin{equation*}
	\begin{split}
    	\log Z_i 
        &= \log \int \mathcal{Q}^{\backslash i}(s_2)\mathcal{P}(u|s_2,v) d_{s_2}d_{v} \int \mathcal{Q}^{\backslash i}(s_1)\mathcal{P}(v|s_1,X) d_{s_1}.
    \end{split}
\end{equation*}
According to the property of conditional Gaussian (Damianou, 2015), we have $\mathcal{P}(v|s_1,X)=\mathcal{N}(v|A_v,B_v)$ where
\begin{equation*}
	\begin{split}
    	A_v &= \mathcal{K}_{hs_1}\mathcal{K}_{s_1s_1}^{-1}s_1,\\
        B_v &= \mathcal{K}_{hh} - \mathcal{K}_{hs_1}\mathcal{K}_{s_1s_1}^{-1}\mathcal{K}_{s_1h} + \sigma_h^2I,
    \end{split}
\end{equation*}
and the form $\mathcal{K}_{ij}$ refers to the kernel matrix. Then the term $\int \mathcal{Q}^{\backslash i}(s_1)\mathcal{P}(v|s_1,X) d_{s_1}$ can be computed by marginalizing out $s_1$ using marginalization and conditioning of Gaussian distributions (Damianou, 2015). The term $\mathcal{P}(u|s_2,v)$ can be defined in the same way as $\mathcal{P}(v|s_1,X)$. After marginalizing out the inducing points $s$, Bui et al. (2016) approximate $$Z \approx \mathcal{N}(u|\mu_u,\Sigma_u),$$ by marginalizing out the hidden layer $v$ where $\mu_u$ and $\Sigma_u$ have explicit formulas. Those formulas involve convolutions of covariance functions with Gaussian distributions. For the general formulas, we refer to Titsias and Lawrence (2010).



\section{Models and Algorithms}
\subsection{Problem Setting}
\label{subsec1}

\vspace{5mm}
We start by formally defining the discrete choice GP (DC-GP) and discrete choice DGP (DC-DGP) models considered herein. Let $F()$ be the joint cumulative distribution function of a multivariate extreme random variable. Let $\mathcal{GP}()$ be a zero mean Gaussian prior and let $\mathcal{K}(\cdot,\cdot)$ be the covariance matrix computed by a predetermined kernel function. The DC-GP model reads
\begin{equation*}
\begin{split}
U_i&=u_i+\epsilon_i^u,\\
(\epsilon_1^u,\epsilon_2^u,...,\epsilon_n^u) &\sim F(\epsilon_1^u,\epsilon_2^u,...,\epsilon_n^u|\lambda_1,\lambda_2,...,\lambda_m),\\
u() &\sim \mathcal{GP}(0,\mathcal{K}_u(\cdot,\cdot)).
\end{split}
\end{equation*}
For simplicity we use $u_i$ to represent $u(x_i)$ in the DC-GP model. In $F()$, $\lambda_m$ is the scale parameter for nest $m$ in a Gumbel distribution. The difference between DC-GP and standard GP is in the noise which is Gumbel distributed instead of Gaussian. This has two consequences: (1) the distribution $\mathcal{P}(D|u)$ is much more complex, and (2) the acquisition function should capture this distribution.

\vspace{5mm}
In the definition of the DC-DGP model, we assume there is one hidden layer $v \in \mathbb{R}^{n\times d}$ and $d$ hidden mappings. It is easy to extend the hierarchy to deeper layers. Each hidden mapping $h^j()$ follows a GP and maps $\{x_i\}_i$ to $\{v^j_i\}_i$. There is also a mapping $u()$ that transforms $\{v_i\}_i$ to the response layer. Each $v_i$ represents a latent feature vector $(v^1_i,v^2_i,...,v^d_i)$. Similar to $u$, we write $v_i$ to represent $v(x_i)$. The network structure is depicted in Figure 2. 
\begin{figure}
 \centering
  \includegraphics[width=130mm]{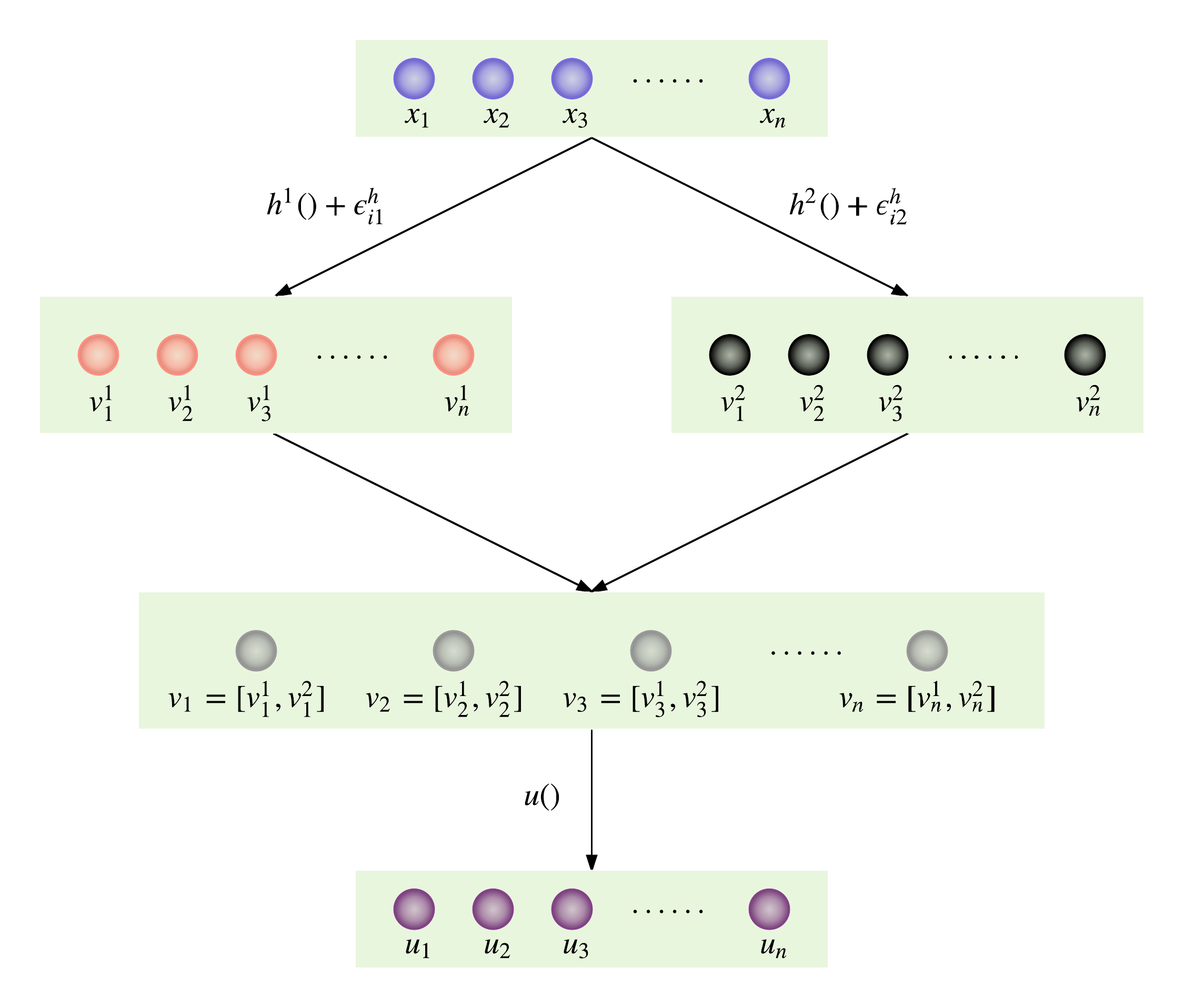}
  \caption{DC-DGP structure with d=2 for 1 hidden layer}
\end{figure}
The model reads
\begin{equation*}
\begin{split}
U_i&=u_i+\epsilon_i^u,\\
v^j_i&=h^j(x_i)+\epsilon_{ij}^h,\\
(\epsilon_1^u,\epsilon_2^u,...,\epsilon_n^u) &\sim F(\epsilon_1^u,\epsilon_2^u,...,\epsilon_n^u|\lambda_1,\lambda_2,...,\lambda_m),\\
\epsilon_{ij}^h &\sim \mathcal{N}(0,\sigma_h^2),\\
u()  &\sim \mathcal{GP}(0,\mathcal{K}_u(\cdot,\cdot)),\\
h^j()  &\sim \mathcal{GP}(0,\mathcal{K}_h(\cdot,\cdot)).
\end{split}
\end{equation*}
Random variable $\epsilon_{ij}^h$ is a noise term and is normally distributed. Here we assume that $\mathcal{K}_u(\cdot,\cdot)$ takes $v$ as input and $\mathcal{K}_v(\cdot,\cdot)$ takes $h$ as input, and thus we have
$$\mathcal{K}_u(\cdot,\cdot)=\mathcal{K}_v(\cdot,\cdot)+\sigma_h^2I,$$
since we assume $\epsilon^hs$ are additive independent identically distributed Gaussian noises (Rasmussen and Williams, 2006). 
\vspace{5mm}

The DC-DGP model has the same differentiating factors as GP. We do want to point out that deep GPs have never been applied before for preference learning.

\vspace{5mm}
\subsection{Algorithm}
In both cases, we follow the algorithm from Section 3.2. In this section we focus on aspects that are different: an approximate preference chain to compute $\mathcal{P}(D|u)$, a novel acquisition function, and algorithmic specifications.

\vspace{5mm}
\subsubsection{DC-GP and DC-DGP algorithms}
In the DC-GP model, we use Laplace approximation to approximate $P(u|D)$ as a function of $u$. To get the gradients and the Hessian matrix, we use $P(D|u)$ and $P(u)$ as explained in Section 3.2. The probability $P(D|u)$ is developed based on the observed pairwise comparisons where we use a preference chain explained later instead of independent pairwise probabilities to approximate it. The distribution $P(u)$ is defined in (4). The model parameter vector $\Theta$ in DC-GP includes kernel and nest parameters.

\vspace{5mm}
In the DC-DGP model, Laplace approximation is also used for the distribution $P(u|D)$ and the preference chain behind the probability $P(D|u)$ is formulated identically as in the DC-GP model. However, due to the complex structure of deep GP models we use $\mathcal{J}(\Theta)$ as explained in Section 3.3 to approximate the distribution $P(u)$. In the DC-DGP model, we find $u^*$ by maximizing the log-likelihood function $\log \mathcal{L} \approx \log P(D|u) + \mathcal{J}(\Theta)$. The model parameter vector $\Theta$ in DC-DGP includes kernel, nest and DGP parameters introduced in (5).

\vspace{5mm}
\subsubsection{Preference chain}
We focus here on estimating $\mathcal{P}(D|u)$. The challenge is the fact that the distribution does not decompose by pairwise comparison and indeed, it seems impossible to get it exactly. 
\vspace{5mm}

We view $D$ as an acyclic graph. We call a preference chain any digraph consisting of a directed path with nodes on the path having at most one other incoming arc with the tail node of such an arc having zero indegree. In layman words, a preference chain has one ``main path" and ``offsprings" of single arcs. In the overall algorithm, we always maintain a preference chain $P$ based on the digraph with respect to $D$. Probability $P(D|u)$ is first approximated by $P(P|u)$ which is further approximated.

\vspace{5mm}

All previous work assume pairwise comparisons are independent but we use the preference chain for two reasons: (1) the preference chain has a better representation of an individual's ordered preference on choices; (2) in our models, Gumbel errors are correlated so the independent assumption does not apply. 

\vspace{5mm}

Probability $P(D|u)$ needs to be computed in each iteration of the algorithm presented in Section 3.2. There are two updating step types: 1) in a MAP optimization iteration, $D$ does not change, but $u$ changes, and 2) from $D_t$ to $D_{t+1}$ one additional comparison is added. During MAP optimization, all model parameters are updated: $u$, $\lambda$, etc.

\vspace{5mm}
At the beginning, we construct the initial $P$ based on the initial $D$ as follows. From each nest we select a random arc in $D$. The highest utility instance (head of the arc) becomes a candidate of $P$ as the main chain. All these instances are nodes in $D$ and we find the longest path among them. This becomes the main path in $P$. To it we add offsprings corresponding to the lower utility instances. The only exception is the lower utility instance from the lowest utility node in the main path which is actually added as a main path node (instead of an offspring). This procedure yields the initial preference chain $P$. In Figure 3, both red and black arcs correspond to $D$ where the connected red arcs represent the main path in $P$ and the dotted arcs stand for offsprings. Note that nest 3 does not have a representative in $P$.

\vspace{5mm}
When moving from $t$ to $t+1$, a new pairwise comparison is added. This pairwise comparison is always between an instance not in $D$ and the highest utility instance $q$ in $P$, i.e. the zero outdegree node in $P$. Based on the comparison in the new pair, the new instance is either added as an offspring to $q$ or to the main path in $P$ as the new highest utility instance. Clearly this preserves the structure of the preference chain.

\vspace{5mm}

We have observed that the initial $P$ has to include many instances from different nests as otherwise $\lambda$s do not update well. For this reason we have designed the appropriate initialization strategy outlined above. 


\begin{figure}
 \centering
  \includegraphics[width=150mm]{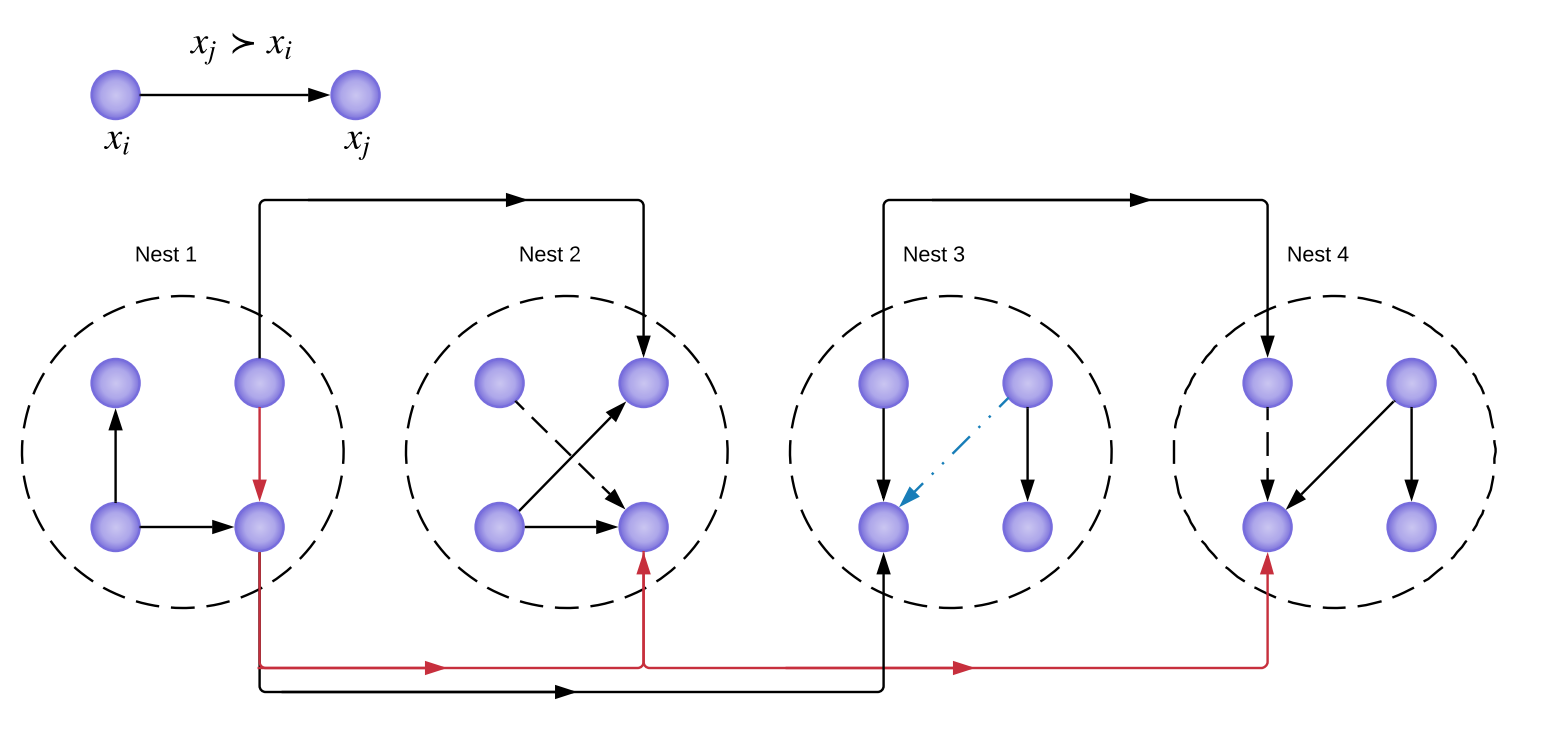}
  \caption{Initial preference chain}
\end{figure}

\vspace{5mm}
Based on Section 3.1, we can derive the probability when the length of a chain is less than or equal to 3. However, the probability of the longest chain for nested logit model is difficult to compute when the length of a chain is greater than 3 using the same method detailed in Dagsvik and Liu (2006). So we further approximate $P(P|u) \approx P(P_1|u)P(P_2|u)$, where $P_1$ is the main path in $P$ and $P_2$ are the offsprings.

\vspace{5mm}
Probability $P(P_2|u)$ is decomposed by pair and each individual probability is computed by (2) or (3). For $P(P_1|u)$ we approximate it by independent triplets (and a possible pair). For triplets we use Table 1. For example,
\begin{equation*}
\begin{split}
\mathcal{P}(x_1 \succ x_2 \succ x_3 \succ x_4 \succ x_5 \succ x_6 \succ x_7 \succ x_8) \\
\approx \mathcal{P}(x_1 \succ x_2 \succ x_3)\mathcal{P}(x_4 \succ x_5 \succ x_6) \mathcal{P}(x_7 \succ x_8).
\end{split}
\end{equation*}


\vspace{5mm}
\subsubsection{Acquisition function with probability improvement}
In Bayesian optimization, acquisition functions are used to sample the next point. Typically, acquisition functions are defined such that high acquisition corresponds to potentially high values of the objective function, whether because the probability of improvement is high, the expected value is high, the uncertainty is low, or both (Brochu, 2010). The probability of improvement is a popular choice. For example, in a noise-free situation (i.e. no Gumbel terms), we can use $$P(U_i>U_j)=\Phi(\frac{\mu_i-\mu_j}{\sigma_i}),$$
with $i$ defined as the index of the next instance to sample, $j$ as the index of the current most preferred instance, and $u_k \sim \mathcal{N}(\mu_k,\sigma_k)$. 
\vspace{5mm}

In both DC-GP and DC-DGP models, the noise is Gumbel distributed and thus we need to compute the probability $\mathcal{P}(u_i+\epsilon_i^u > u_j+\epsilon_j^u)$. 
\vspace{5mm}

Let $f(x;\alpha)$ define the logistic distribution with location 0 and scalar $\alpha$. We have $P(U_i>U_j)=P(u_i+\epsilon_i^u > u_j+\epsilon_j^u)=P(\delta_j-\delta_i+\epsilon_j^u-\epsilon_i^u<\mu_i-\mu_j)$ where $\delta_k\sim \mathcal{N}(0,\sigma_k)$ and because $u$ is based on GP.

\vspace{5mm}
According to Crooks (2009), we can approximate $\delta_j-\delta_i+\epsilon_j^u-\epsilon_i^u$ as a reparameterized logistic function
\begin{equation*}
\delta_j-\delta_i+\epsilon_j^u-\epsilon_i^u \sim f(x;\gamma\lambda_m), \quad \gamma=\sqrt{1+\frac{\pi(\sigma_i^2+\sigma_j^2)}{8\lambda_m^2}}.
\end{equation*}

Here $\lambda_m=1$ if $i,j$ are in different nests and otherwise $\lambda_m \in (0,1)$ according to the common nest $m$.
\vspace{5mm}

To summarize, we have the probability improvement function or acquisition function defined as
\begin{equation}
P(U_i>U_j)=\frac{1}{1+e^{\frac{-(\mu_i-\mu_j)}{\gamma\lambda_m}}}.
\end{equation}

\vspace{5mm}
\subsubsection{Active learning algorithm}

The general framework of the active learning algorithm is presented in Algorithms 1 and 2. In Algorithm 1, $X^u$ refers to instances which have not yet been compared and $X^l$ is a set containing instances which have already been compared, i.e. they appear at least once in $D$. Value $xBest$ represents the instance yielding the highest utility up to iteration $t$. Hyperparameter $\zeta$ is the lower threshold on probability improvement. The algorithm covers a single step of going from $t$ to $t+1$ (index $t$ is omitted for clarity). 
\vspace{5mm}

Function $preferenceChain$ takes $D$ and returns preference chain $T$ based on the method outlined in Section 4.2.2. Function $optimizer$ based on Section 3.2 returns the Hessian matrix $\mathcal{H}$ computed at MAP $u^*$ with corresponding $x^*$, estimated covariance matrix $\Sigma$, maximum estimated mean $\mu_{max}$ of all alternatives in $X^l$ and model parameters $\Theta$. Function $evaluation$ returns the acquisition value. Function $oracle$ returns the true preference comparison between two given choices. In Algorithm 2, the probability improvement is computed. Index $im$ refers to the index of $x^*$ in $\Sigma$. Function $Predict$ is based on the predictive distribution in Brochu et al. (2007). We use $\Phi_g$ to compute the cumulative density of the normal-Gumbel convolution based on (6).

\vspace{5mm}
\begin{algorithm} [H]
\SetAlgoNoLine
\textbf{inputs}: $D$, $xBest$\\
$T \gets preferenceChain(D)$\\
$(\mathcal{H},\Sigma,\Theta,x^*,u^*,\mu_{max})=optimizer(T,D)$\\
$x^l \gets \emptyset$\\
$PI^*=\zeta$ \\ 

	

\ForEach{$x_i \in X^u$}{
	
    $PI = evaluation(\mathcal{H},\Sigma,\Theta,\mu_{max},x_i,u^*)$\ \\ 

\uIf{$PI \geq PI^*$}{
$PI^*=PI$\\
$x^l \gets x_i$\\
}}


\uIf{$x^l \; equals \; xBest$}{
return $True$\\
}
\Else{

\uIf{$oracle(x^l,x^*)$ is $true$}{

Update $D, X^l$ with new relation $x^l \succ x^*$\\

}\Else{
Update $D, X^l$ with new relation $x^* \succ x^l$\\
}
Remove $x^l$ from $X^u$\\
return $False$
}

\caption{Active learning approach for a given $t$}
\end{algorithm}

\vspace{5mm}
\begin{algorithm} [H]
\SetAlgoNoLine
\textbf{inputs}: $\mathcal{H}$, $\Sigma$, $\Theta$, $\mu_{max}$,$x_i$, $u^*$\\
$\sigma^* \gets \Sigma_{im,im}$\\
$(u_i,\sigma_i) \gets Predict(\mathcal{H}, \Sigma, \Theta, x_i)$\\
$PI=\Phi_g(u_i,u_{max},\sigma^*,\sigma_i,\Theta)$\\

\textbf{output: $PI$}

\caption{$evaluation()$}
\end{algorithm}

\label{subsec1}

\section{Experiments}
We implement the proposed DC-GP and DC-DGP models in two different scenarios. The first scenario is to measure capabilities of the algorithm of searching a maximum of some latent functions. The latent functions come from a prior work (Brochu et al., 2007), but we modify them by adding correlated Gumbel noises to the function values. The second scenario applies the models to a public airline itinerary dataset. We implement the algorithms in Python 2.7 and run them on a Linux server with Intel Xeon CPU at 2.2 GHZ and 32 GB memory. In training, to find the hyperparameters in the kernel function and NL parameters $\lambda_m$, we optimize them together with hidden $u$ by using Adam as the optimization algorithm in both DC-GP and DC-DGP models. 

\vspace{5mm}
For each experiment, we create 10 different starting scenarios where each scenario starts with different pairwise comparisons. As the algorithm requires computing nest parameters, to ensure the starting scenarios include instances from every nest we use a two-phase comparison method. In the first phase, we randomly select two instances from each nest and compare them. In the second phase, we use the preferred instances (i.e. the ``winners" in the comparisons) from the first phase and compare them to get a complete order among them. We use two different DC-DGP model structures: one model with a one dimension hidden layer named as DC-DGP1 and one model with a five dimension hidden layer named as DC-DGP5. The results that follow represent overall performance values across these 10 random starting scenarios.

\vspace{5mm}
To benchmark the active learning algorithm, we create a random search method. At each query $t$, it randomly selects an instance from those that have not yet been compared. The random search method applies to the same starting scenarios as our algorithm. To obtain an average performance, we run it 500 times for each scenario. With 500 runs, the standard deviation of performance values is low.

\subsection{Latent Function Experiments}
\subsubsection{Experiment Design}
The functions we evaluate are defined as
\begin{equation*}
	\begin{split}
    	f_{2D} &= max\{0,-1+\sum_{i=1}^2 sin(x_i)+x_i/3+sin(12x_i)\},\\
        f_{4D} &= \sum_{i=1}^{4} sin(x_i)+x_i/3+sin(12x_i),\\
        f_{6D} &= \sum_{i=1}^{6} sin(x_i)+x_i/3+sin(12x_i),\\
        x_i &\in [0,1],
    \end{split}
\end{equation*}
which come from Brochu et al. (2007).
\vspace{5mm}

To attach correlated Gumbel noises to $f_{2D},f_{4D}$ and $f_{6D}$ values, we first discretize the space which also provides an effective reflection to our active learning problem setting (i.e. a finite number of instances). In discretization, we split each dimension equally to obtain a grid. In different latent functions we choose a different grid size. In the 2D function, we use 0.05 as the discretization length so the dataset has $22 \times 22 = 484$ points. In the 4D function, we use 0.2 and have 1,296 points. In the 6D function, we use 0.25 which yields 15,625 points. 
\vspace{5mm}

Now we explain how to allocate each point to different nests. Based on the contour plot of the 2D function, there are 4 local maxima at points (0.15, 0.15), (0.15, 0.65), (0.65, 0.15), (0.65, 0.65). We assume that each local maximum represents a nest. To decide the nest for other points, we use these four centers and assign other grid points based on their Euclidean distances to the centers. In the 2D case, we end up with 4 nests. In the 4D and 6D case, as the functions are symmetric it is easy to infer their local maxima (i.e. the centers). However in the 4D case, this strategy creates a large number of 16 nests which is not typical in NL model studies. We combine them as follows. One center in the 4D case is (0.15, 0.15, 0.15, 0.65). Since the function is symmetric, some permutations of these coordinates are local maxima. We assume that all such permutations are in the same nest. As the functions are symmetric, this strategy guarantees that local maxima with the same value belong to the same nest. In this way, we divide all points into 5 nests in the 4D case. To consolidate nests for the 6D case, we use the same method as in the 4D case to obtain 7 nests.
\vspace{5mm}

After space discretization and nest assignment, reasonable noises are generated. In most empirical studies within the transportation industry, the nest parameter $\lambda_m \in [0.5, 1)$, hence we use this fact to derive parameters to generate Gumbel noises. Firstly, we rank the nests based on the objective value of the center. Then we generate a mean value for each $\lambda_m$ as follows. In the 4D function, we have 5 nests, and we assume the mean values $\mu_\lambda$ are taken sequentially from (0.80, 0.75, 0.70, 0.65, 0.60). The variances are configured based on coefficient of variation $\sigma_\lambda/\mu_\lambda=0.5$. We then use a truncated normal distribution to sample $\lambda_m$ with lower and upper bounds to be $\mu_{\lambda}\pm 0.05$. Taking 0.60 as an example, we use $\mathcal{N}(\lambda|\mu=0.60,\sigma=1.2,a=0.55,b=0.65)$.

\subsubsection{Results}
The results from running the algorithm on the 2D/4D/6D functions are presented in Figures 4 to 6. The x-axis represents the relative gap between the current best found value (i.e. at query step $t$) and the maximum of the latent function. The y-axis is the number of queries (or additional pairwise comparisons) the algorithm requests. The lower a number of queries is, the better.

\vspace{5mm}
In general, the DC-DGP models outperform the DC-GP model and the random method after the $5^{th}$ query in the 2D experiment. In contrast, the DC-GP model is only better than the random method after the $40^{th}$ query. In the 4D experiment in Figure 5, we observe a similar trend as in the 2D case. However, in the 6D experiment all three models outperform the benchmark quickly. In summary,
\begin{itemize}
	\item in the 2D experiment, to achieve the same performance of the benchmark at the $50^{th}$ query, the DC-DGP models take around 10-12 queries;
    \item in the 4D experiment, to achieve the same performance of the benchmark at the $50^{th}$ query, the DC-DGP models take around 20 queries; and
    \item in the 6D experiment, to achieve the same performance of the benchmark at the $50^{th}$ query, the DC-DGP models take around 18-21 queries.
\end{itemize}
As the dimensionality is increasing, the performances of the DC-DGP models are more pronounced. Secondly, when the dimensionality increases and the number of instances increases sharply, the DC-GP model has a performance boost. It is not only better than the benchmark but also yields a comparable performance as the DC-DGP models within the first 10 queries. This indicates that when the latent function's complexity is increasing, the DC-GP model's performance increases. We think at low dimensionality, the DC-GP model using the proposed acquisition function is easily trapped at a local optimum so its performance is worse than the random method in the 2D and 4D cases. As for DC-DGP1 and DC-DGP5, when the function's complexity increases, the more complex model (i.e. DC-DGP5) explores a better representation of the hidden function and yields a better result.

\vspace{5mm}
In Table 3, we present the computation time comparison in the 6D case for the first 10 queries based on our proposed acquisition function.

\vspace{5mm}
The main conclusions are as follows.
\begin{itemize}
	\item There is not much of a difference in the first 5 queries. The random method is worse but not by much.
    \item After 5 queries, DC-DGP models drastically outperform the random method.
    \item In a low dimension, DC-DGP1 performs better than DC-DGP5. As complexity increases, DC-DGP5 becomes better.
    \item As the number of queries increases, DC-DGP5 is much better.
    \item The DC-GP model with the proposed acquisition function does not work well for small problems.
\end{itemize}

In summary, if a problem is high dimensional with a large number of choices, we recommend DC-DGP5. If the dimensionality is low, we should use DC-DGP1. If a medium or low number of queries is required, we advise to use DC-DGP1, else DC-DGP5.

\vspace{5mm}

Furthermore, we also test the adaptive upper confidence bound (UCB) (Brochu, 2009) function. This function is defined as

\begin{equation*}
	u(x) + \sqrt{\tau_t}\sigma(x),
\end{equation*}
\begin{equation*}
	\tau_t = 2\log(t^{p/2+2}\pi^2/3\delta),
\end{equation*}

where $t$ is the $t^{th}$ step, $p$ is the dimension of $x$ (e.g. 6 in the 6D case), and $\delta$ is a random value drawn from the uniform distribution between 0 and 1. We have results using adaptive UCB compared with our proposed acquisition function in the 2D, 4D and 6D experiments in Appendix 8.1, Figures 9 to 11. From the comparisons, we draw three main conclusions as follows.
\begin{itemize}
	\item The DC-GP model with adaptive UCB works better than our proposed acquisition function and it outperforms the benchmark.
    \item The DC-DGP models with adaptive UCB are not as effective as our proposed acquisition function when a problem is low dimensional.
    \item The DC-DGP models with adaptive UCB can be more effective than our acquisition function if a problem is high dimensional.
\end{itemize}

\begin{figure}
 \centering
  \includegraphics[width=160mm]{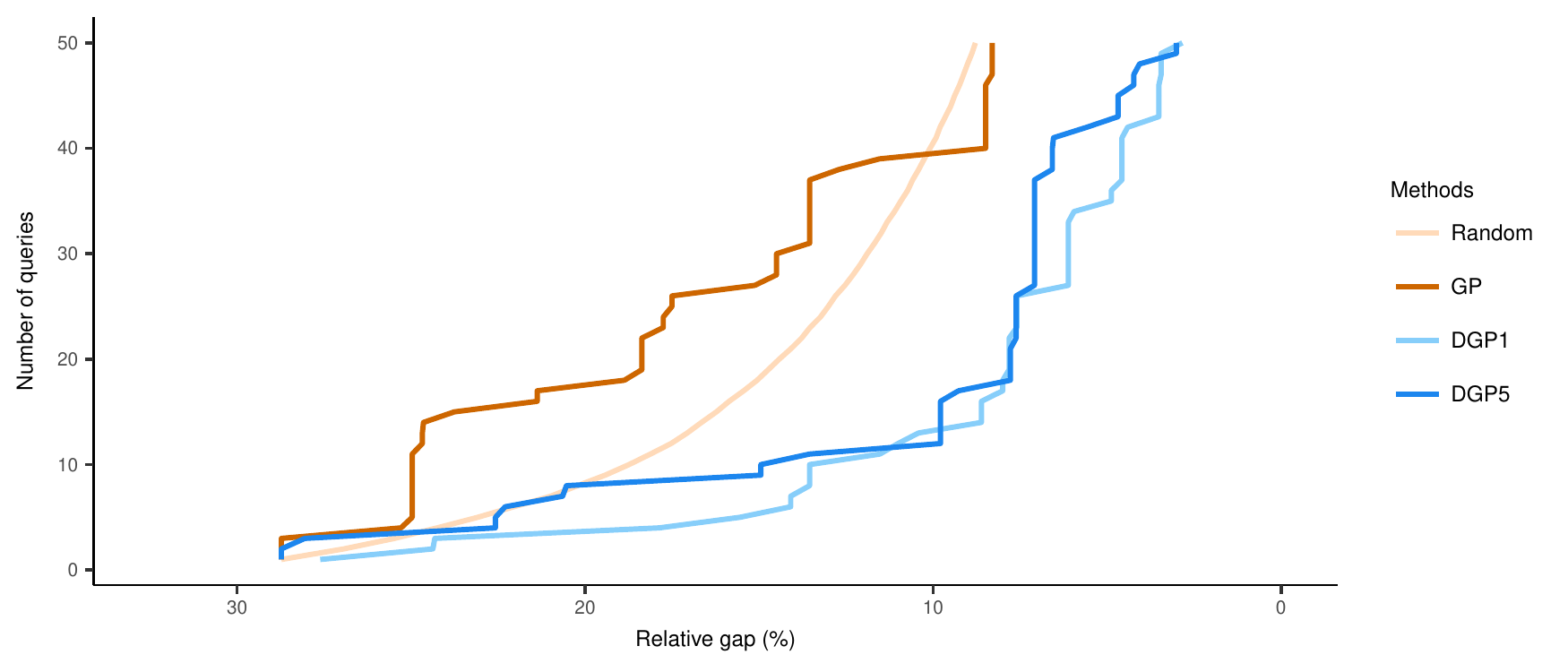}
  \caption{Search queries for 2D function, N=484, number of repetitions = 10}
\end{figure}

\begin{figure}
 \centering
  \includegraphics[width=160mm]{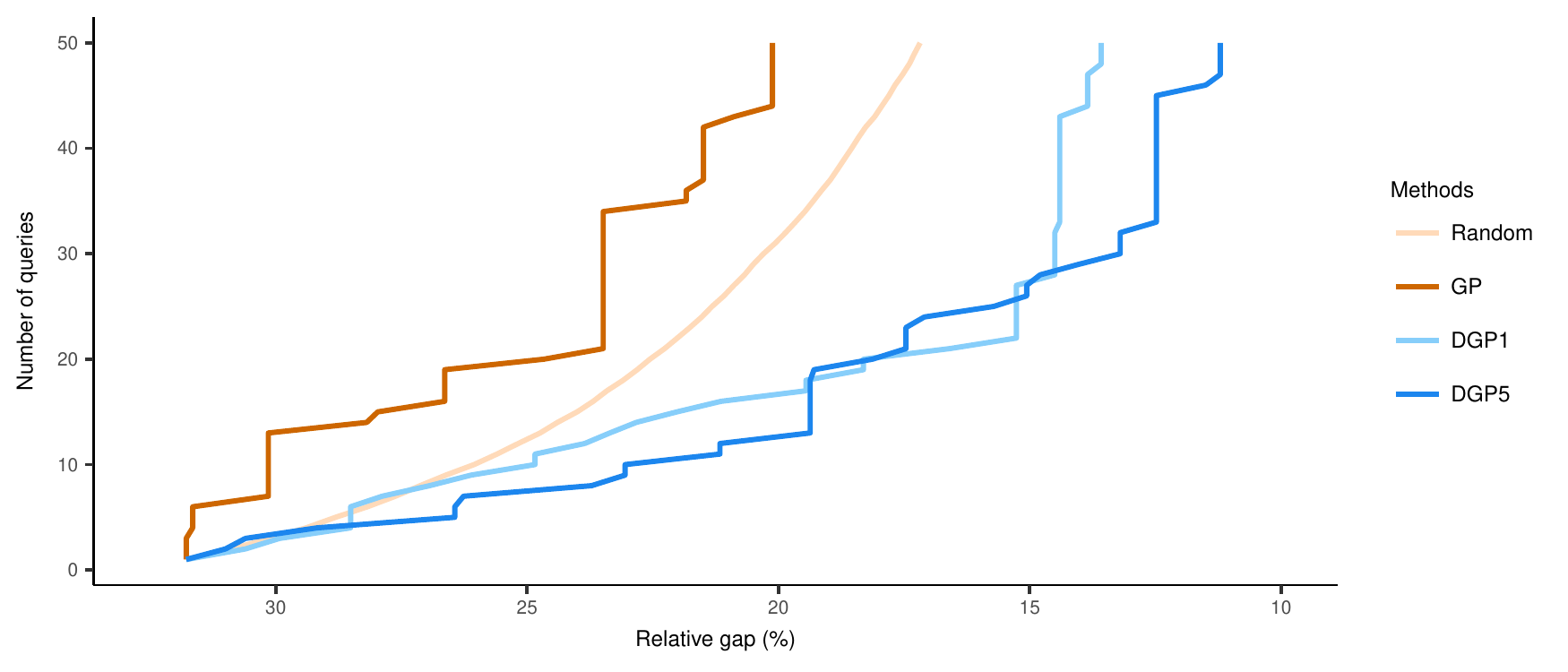}
  \caption{Search queries for 4D function, N=1,296, number of repetitions = 10}
\end{figure}
\vspace{5mm}

\begin{figure}
 \centering
  \includegraphics[width=160mm]{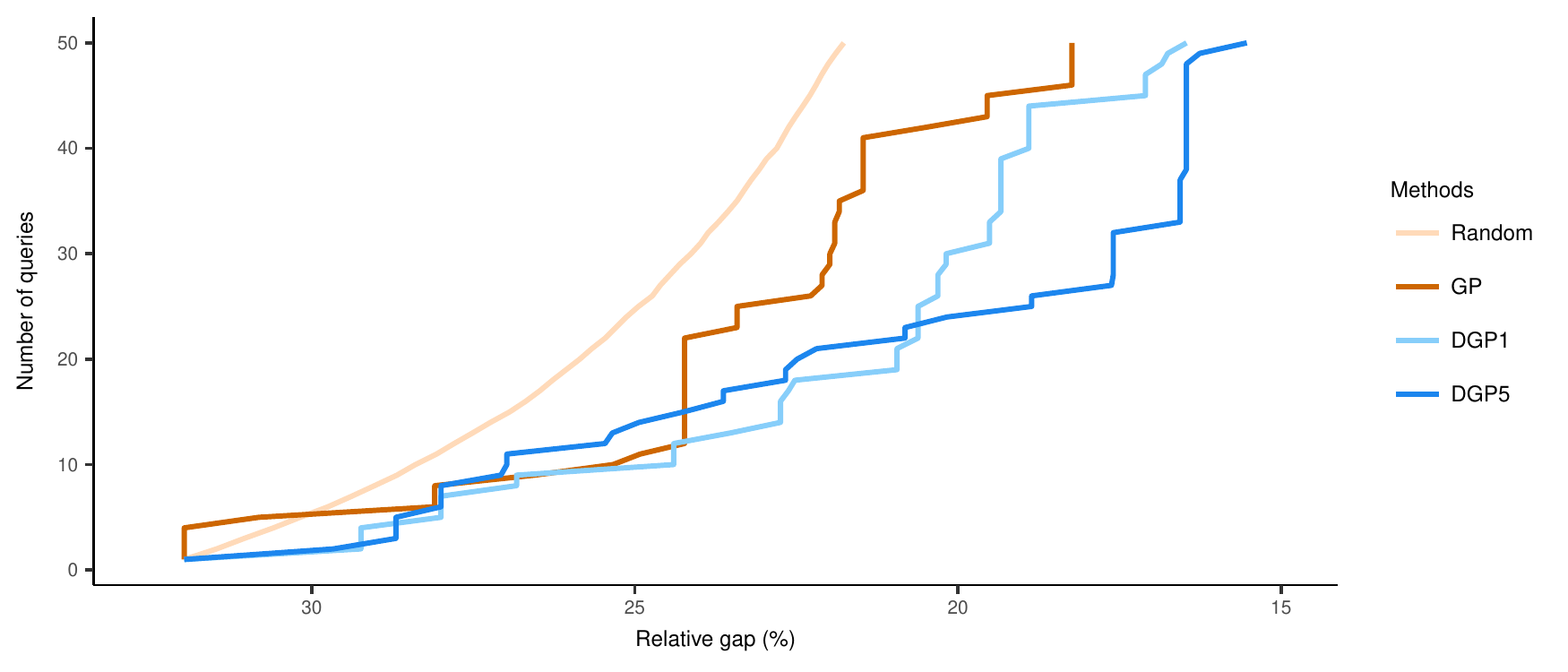}
  \caption{Search queries for 6D function, N=15,625, number of repetitions = 10}
\end{figure}

\subsection{Airline Itinerary Experiment}
The itinerary dataset can be obtained from \url{https://github.com/jpn--/larch}. It is from U.S. continental markets in May of 2013 and is retrieved from a ticketing database provided by the Airlines Reporting Corporation. Each choice set has a booked itinerary (i.e. no rank). To construct the preference rank, we use all choice sets to build an NL model first. We then pick the largest five choice sets and we assume they are from one choice set. This yields in total 543 itineraries. Next the parameters and logsum terms from the NL model are applied to compute the probability of choosing each itinerary.

\subsubsection{An NL itinerary model}
Based on the level of service (i.e. stop and non-stop) and time of day (i.e. morning, afternoon and evening), we divide the itineraries into six different nests. Similar to the classic NL models, we assume the logsum parameters $\lambda$s are the same for all nests. Then the probability of choosing itinerary $i$ belonging to nest $m$ is formulated as $$\mathcal{P}_i^m=\frac{\exp(\lambda\Gamma_m)}{\sum_{m^{'} \in M}\exp(\Gamma_m^{'})}\times \frac{\exp(\frac{1}{\lambda}U_i)}{\sum_{i^{'}\in m}\exp(\frac{1}{\lambda}U_{i^{'}})},$$ where $\Gamma_m=\ln(\sum_{i^{'}\in m}\exp(\frac{1}{\lambda}U_{i^{'}}))$ and $M$ is the set of all nests. The estimated model parameters are summarized in Table 2. Time period 2 to 9 is originally encoded in the dataset which stand for different departure time windows. Equipment type refers to the aircraft type.

\begin{center}
Table 2. NL model parameters
\end{center}
\begin{center}
\begin {tabular}{c | r|c|r}
\hline
Explanatory variables & Parameters &Explanatory variables & Parameters \\
\hline
Time period 2&0.082&Carrier 2 & 0.095\\

Time period 3&0.104&Carrier 3 & 0.500\\

Time period 4&0.069&Carrier 4 & 0.435\\

Time period 5&0.110&Carrier 5 & -0.508\\

Time period 6&0.204&Equipment type 2 & 0.379\\

Time period 7&0.259&High yield fare & -0.001\\

Time period 8&0.265&Low yield fare & -0.001\\

Time period 9&-0.076&Elapsed time & -0.005\\

Non-stop & -3.090&&\\
\hline
Logsum & 0.792 \\
Log-likelihood at convergence&-777501.590\\

Log-likelihood at zero&-953940.440\\

Rho Squared w.r.t. Null Parameters & 0.185\\
\hline
\end {tabular} 
\end{center}
Note: All explanatory variables and the logsum parameter $\lambda$ are significantly different from zero at level 0.01.

\subsubsection{Results}
We first present the results of the itinerary experiment in Figure 7 by using the proposed acquisition function. Note that this is a 17 dimensional problem. The figure shows that the average performances of the DC-GP and DC-DGP models only outperform the benchmark within the first 15 queries. Since the results indicate the proposed models are underperformed after some point, we also test the adaptive UCB. We get the results presented in Figure 8. In the figure, we observe that DC-DGP5 is the most preferred model if the target is to find the best itinerary within 20 queries, otherwise DC-DGP1 should be used.

\vspace{5mm}
\subsection{Summary}
Table 4 shows the performances of the DC-GP and DC-DGP models in all experiments. We then summarize the recommendations as follow.
\begin{itemize}
	\item If the number of instances is high (i.e. 6D) and computation time is a concern (e.g. the DC-DGP1 and DC-DGP5 spend 50\% to 70\% more time than the DC-GP model), we recommend using the DC-GP model with adaptive UCB.
	\item When the feature dimension is high and the running time is not a concern, we recommend a DC-DGP model with more hidden mappings (DC-DGP5) with our proposed acquisition function if it requires a few queries, and less hidden mapping (DC-DGP1) with adaptive UCB if a high number of queries is tolerable.
    \item If the number of instances is low (i.e. 2D) or in a low feature number context and time is a concern, we recommend a DC-GP model using adaptive UCB.
    \item When the number of instances is low (i.e. 2D) or in a low feature number context, we recommend a DC-DGP model with less hidden mappings and our proposed acquisition function if time is not a concern.
\end{itemize}

\begin{figure}
 \centering
  \includegraphics[width=160mm]{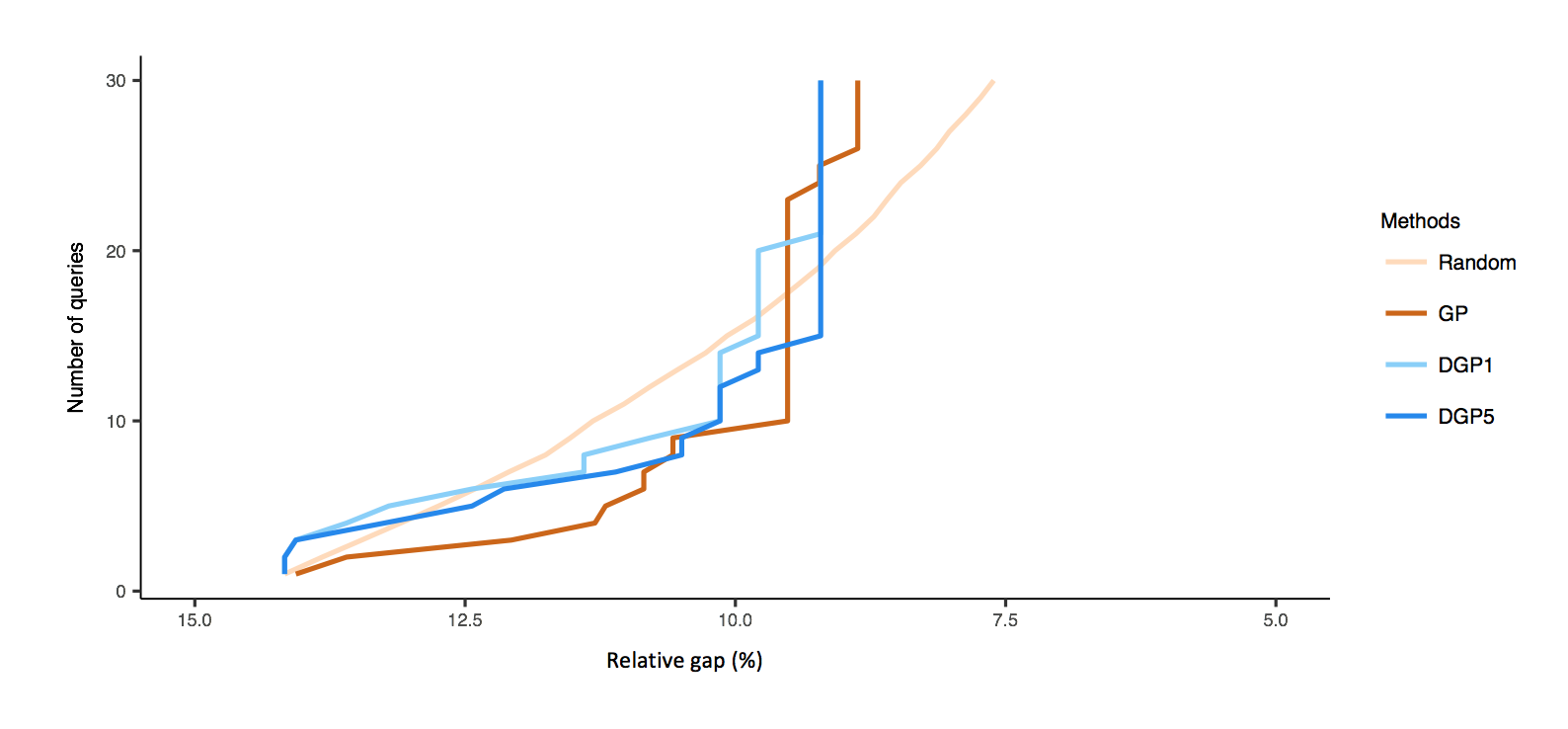}
  \caption{Search queries for airline dataset with proposed acquisition function, N=543, number of repetitions = 10}
\end{figure}

\begin{figure}
 \centering
  \includegraphics[width=160mm]{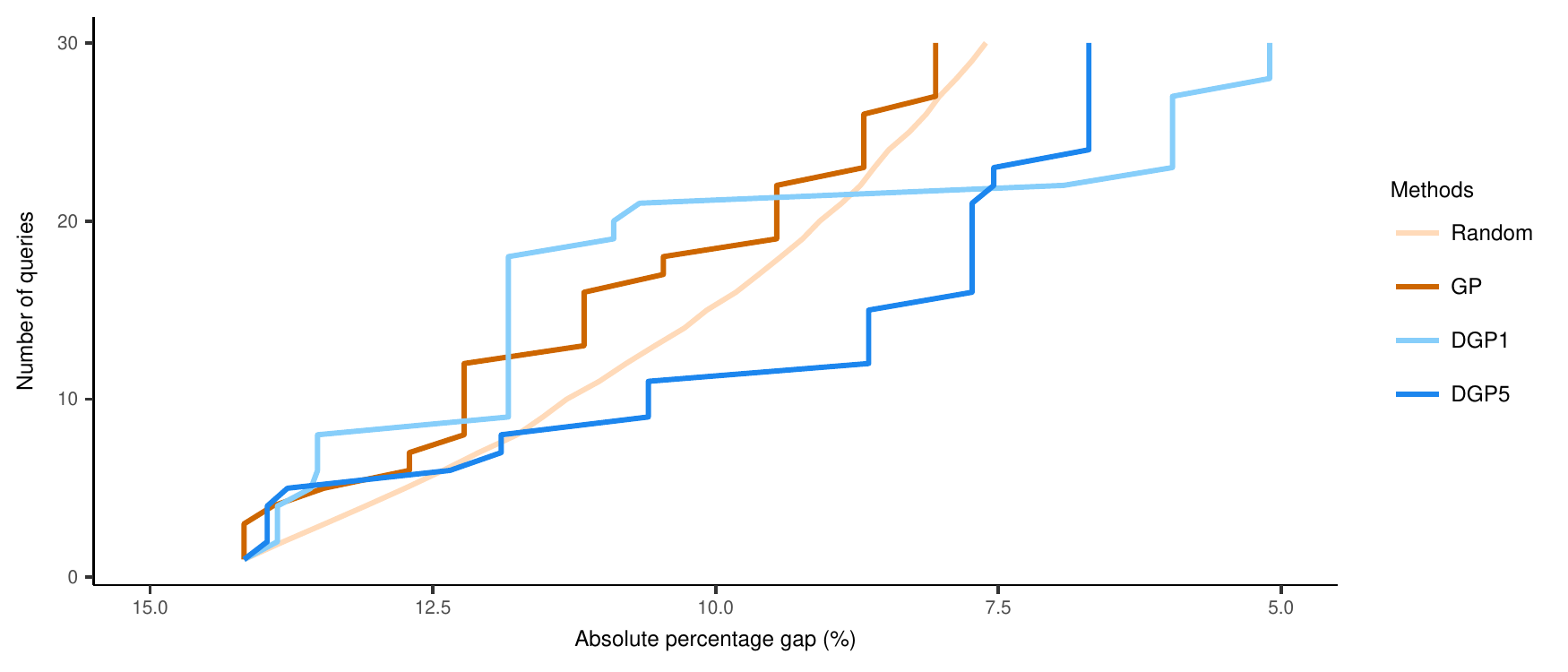}
  \caption{Search queries for airline dataset with adaptive UCB, N=543, number of repetitions = 10}
\end{figure}

\vspace{5mm}

\begin{table}[h!]
\begin{center}
Table 3. Computation time for the first 10 queries in the 6D experiment (in seconds)
\end{center}
\begin{center}
\begin {tabular}{c | c|c}
\hline
GP & DGP-1 & DGP-5 \\
\hline
6,076 & 8,557 (+41\%)& 11,720 (+93\%)\\
\hline
\end {tabular} 
\end{center}
\end{table}

\begin{table}[h!]
\begin{center}
Table 4. Performance comparisons
\end{center}
\begin{center}
\begin {tabular}{c | c|c|c|c}
\hline
Query  \# & 2D & 4D & 6D & Airline\\
\hline
$\leq$ 10 & DC-DGP1&DC-DGP5 & DC-DGP5&DC-DGP5 (UCB)\\

10-20&DC-DGP1&DC-DGP5 &DC-DGP1/DC-DGP5 &DC-DGP5 (UCB)\\

20-30&DC-DGP1&DC-DGP1/DC-DGP5 & DC-DGP5&DC-DGP1 (UCB)\\

30-40&DC-DGP1&DC-DGP5 & DC-DGP1 (UCB)&-\\

40-50&DC-DGP1&DC-DGP5 & DC-DGP1 (UCB)&-\\
\hline
\end {tabular} 
\end{center}
\end{table}

\section{Conclusion}
In this paper, we present a new active preference learning algorithm which embeds GP and DGP models. The models explore a novel way of learning preference from pairwise data which include classic choice model assumptions. A new acquisition function has also been proposed to solve this problem. The new methods have been applied to four different experiments and the proposed DC-DGP models have been proven to be more effective than the DC-GP model and a random search method. The issues of underperformed situations can be explained by the complexity of the latent functions and the choice of acquisition function. In general, we conclude that when the choice set is large and we require fast response, then the DC-GP model is preferred. On the other hand, if computational time is not a big concern, we recommend using the DC-DGP models with complex structure to get a better performance. The proposed method can be applied not only in the airline or travel industry but also in other scenarios involving disentangling customers' preferences in a sequential way. In the future, we will investigate other approximation methods in the DGP models and alternative acquisition functions which may provide good solutions faster.


\section{Appendix}
\subsection{2D, 4D and 6D with adaptive UCB compared to original results}
In Figures 9-11, we show the comparison of the proposed acquisition function and UCB.

\begin{figure}
 \centering
  \includegraphics[width=160mm]{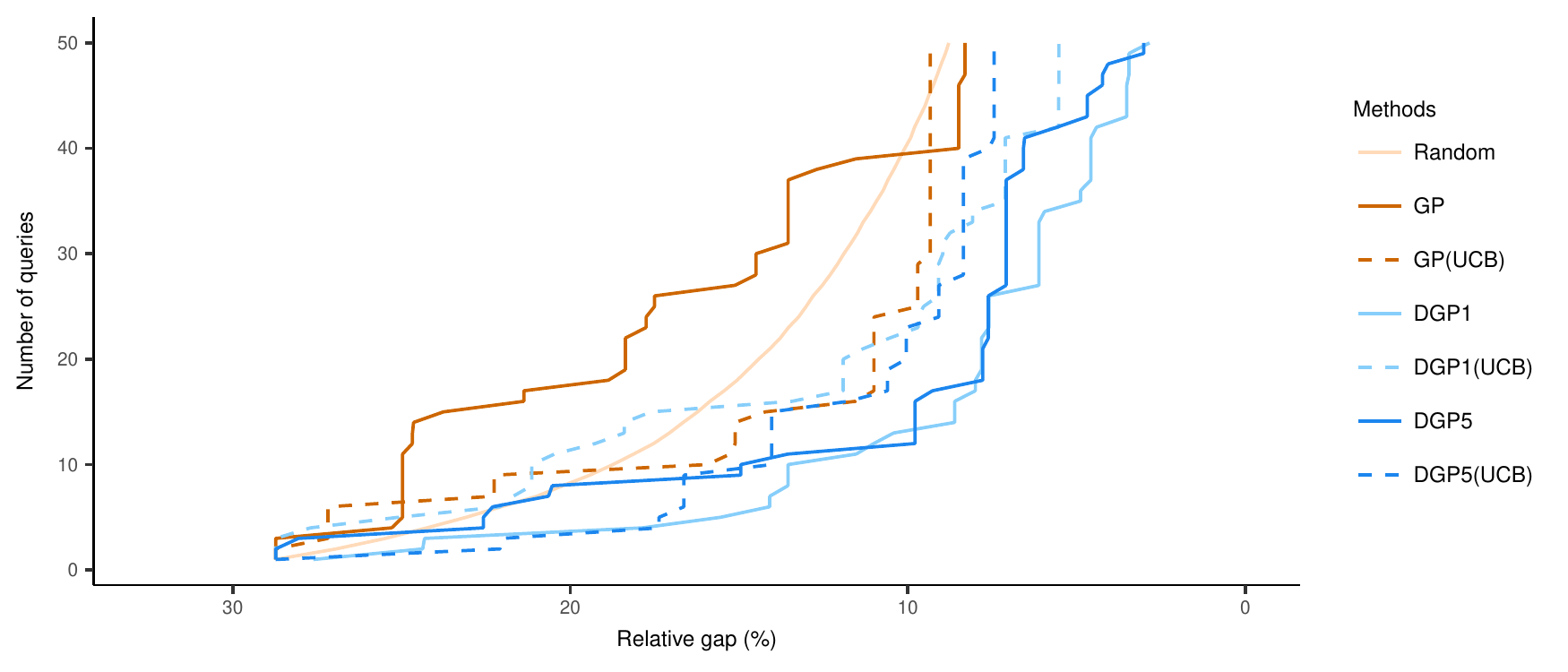}
  \caption{Search queries for 2D function with adaptive UCB compared to original results, N=484, number of repetitions = 10}
\end{figure}

\begin{figure}
 \centering
  \includegraphics[width=160mm]{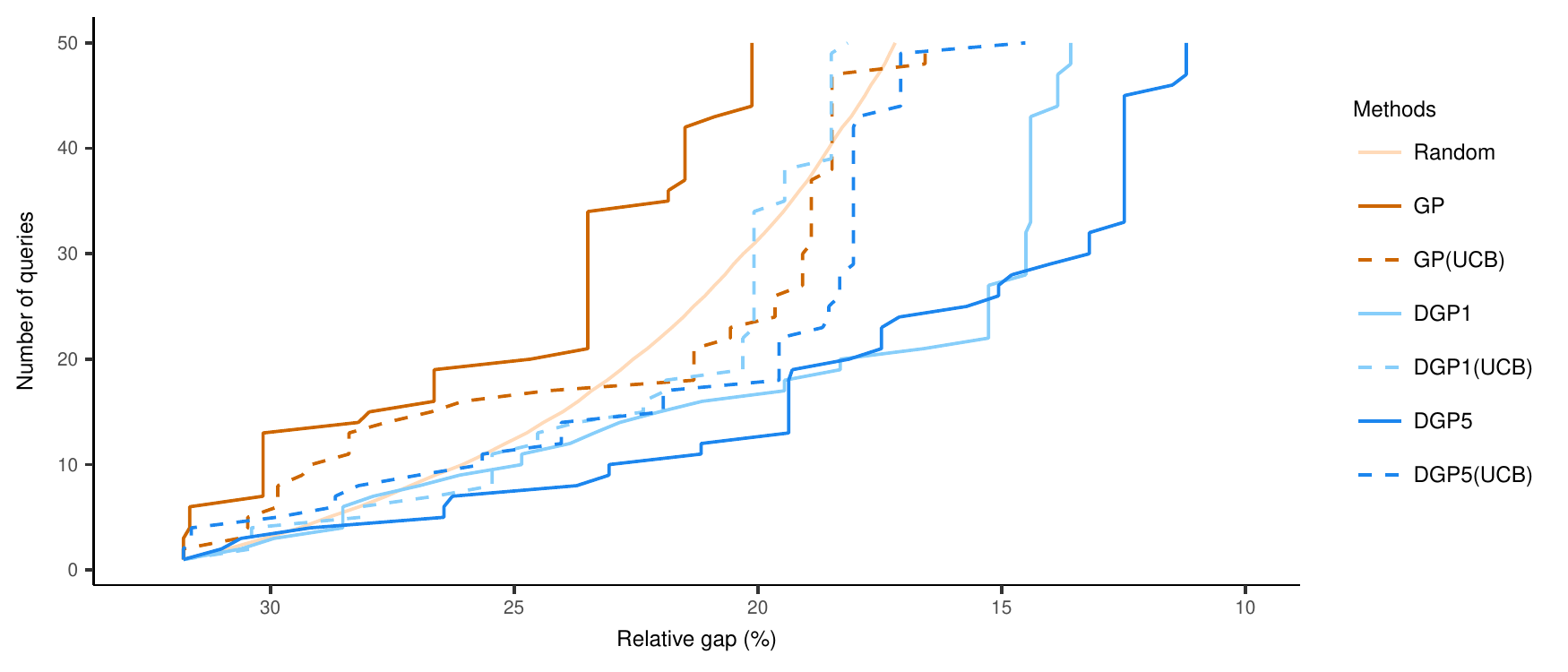}
  \caption{Search queries for 4D function with adaptive UCB compared to original results, N=1,296, number of repetitions = 10}
\end{figure}
\vspace{5mm}

\begin{figure}
 \centering
  \includegraphics[width=160mm]{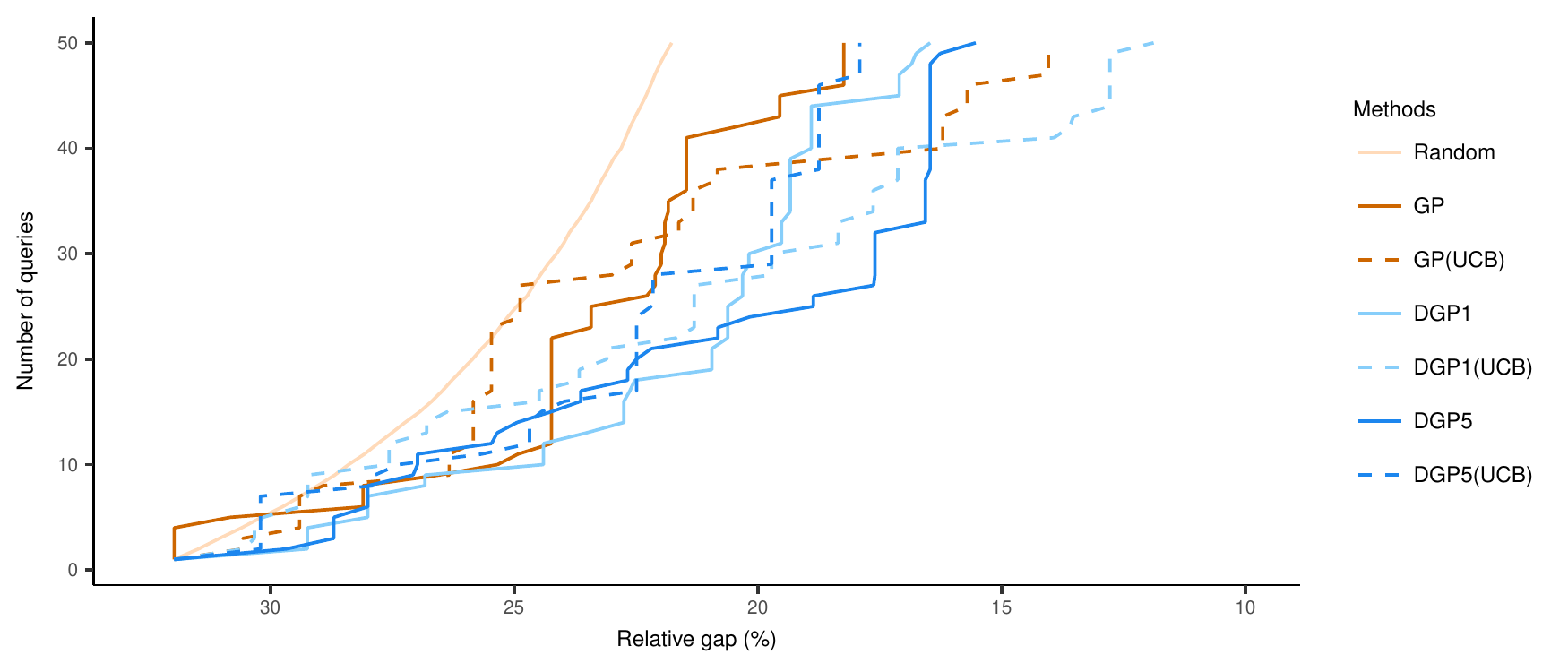}
  \caption{Search queries for 6D function with adaptive UCB compared to original results, N=15,625, number of repetitions = 10}
\end{figure}




\bibliographystyle{elsarticle-num}

\bibliography{sample}

\end{document}